\documentclass[10pt,twocolumn,letterpaper]{article}
\usepackage{cvpr}     
\usepackage{graphicx}

\usepackage{makecell}
\usepackage{colortbl}
\usepackage{multirow}
\usepackage{xcolor}    
\usepackage{arydshln}   

\definecolor{cvprblue}{rgb}{0.21,0.49,0.74}
\usepackage[pagebackref,breaklinks,colorlinks,allcolors=cvprblue]{hyperref}
\def\paperID{} 
\def\confName{CVPR}
\def\confYear{2026}

\title{UniFusion: A Unified Image Fusion Framework with Robust Representation and Source-Aware Preservation}

\author{Xingyuan Li\textsuperscript{\rm 1*}, Songcheng Du\textsuperscript{\rm 2*}, Yang Zou\textsuperscript{\rm 2}, HaoYuan Xu\textsuperscript{\rm 3}, Zhiying Jiang\textsuperscript{\rm 4}\dag, Jinyuan Liu\textsuperscript{\rm 3}\\
\textsuperscript{\rm 1} Zhejiang University, 
\textsuperscript{\rm 2} Northwest Polytechnical University, \\
\textsuperscript{\rm 3} Dalian University of Technology, 
\textsuperscript{\rm 4} Dalian Maritime University \\
{\tt\small xingyuan\_lxy@163.com, dusongcheng@mail.nwpu.edu.cn} \\
}
\begin{document}

\twocolumn[{%
\renewcommand\twocolumn[1][]{#1}%
\maketitle

\begin{center}
    \centering
    \captionsetup{type=figure}
    \vspace{-0.25in}
    \includegraphics[width=1\textwidth]{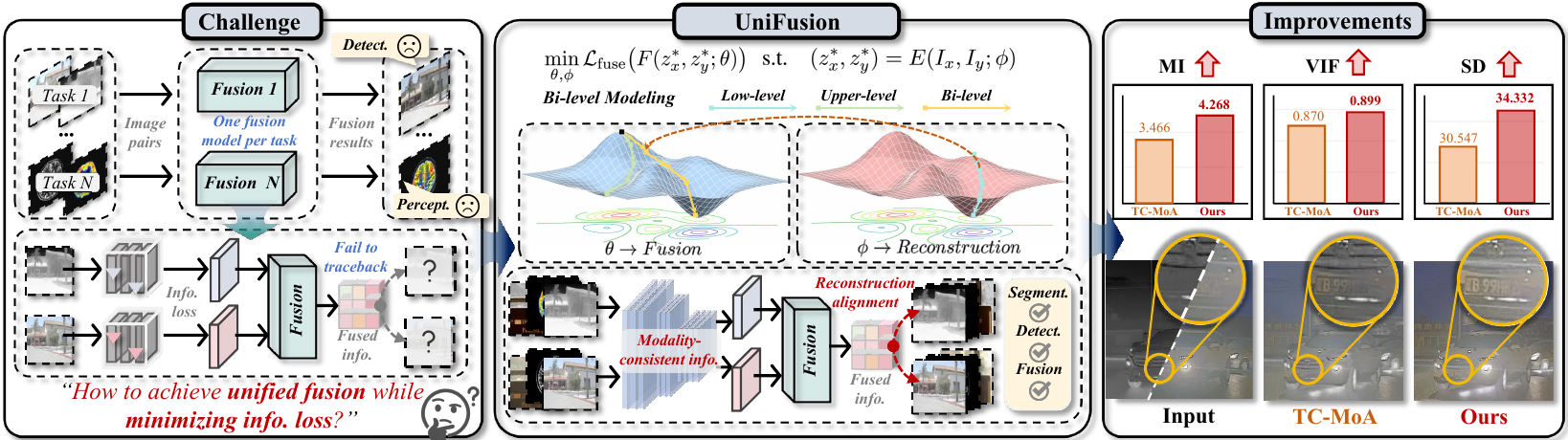}
    \vspace{-0.2in}
    \captionof{figure}{Overview of the proposed bilevel optimization framework for unified image fusion task. \textbf{(Left)} Conventional approaches are typically designed for task-specific scenarios and struggle to effectively preserve source information during the fusion process, leading to inconsistent modality representation and information loss. \textbf{(Middle)} Our UniFusion formulates fusion as a bilevel optimization problem. The lower-level reconstruction branch learns modality-consistent representations through self-reconstruction, while the upper-level fusion branch adaptively integrates them into a unified representation that effectively enhances semantic preservation and structural integrity. \textbf{(Right)} Quantitative and qualitative results demonstrate that our method consistently outperforms TC-MoA across multiple metrics.}
    \label{fig:teaser}
\end{center}%
}]

\renewcommand{\thefootnote}{}
\footnote{*These authors contributed equally.}
\footnote{\dag Corresponding author.}
\renewcommand{\thefootnote}{\arabic{footnote}}

\begin{abstract}
Image fusion aims to integrate complementary information from multiple source images to produce a more informative and visually consistent representation, benefiting both human perception and downstream vision tasks. Despite recent progress, most existing fusion methods are designed for specific tasks (i.e., multi-modal, multi-exposure, or multi-focus fusion) and struggle to effectively preserve source information during the fusion process. This limitation primarily arises from task-specific architectures and the degradation of source information caused by deep-layer propagation. To overcome these issues, we propose \textbf{UniFusion}, a unified image fusion framework designed to achieve cross-task generalization. First, leveraging DINOv3 for modality-consistent feature extraction, UniFusion establishes a shared semantic space for diverse inputs. Second, to preserve the understanding of each source image, we introduce a reconstruction-alignment loss to maintain consistency between fused outputs and inputs. Finally, we employ a bilevel optimization strategy to decouple and jointly optimize reconstruction and fusion objectives, effectively balancing their coupling relationship and ensuring smooth convergence. Extensive experiments across multiple fusion tasks demonstrate UniFusion’s superior visual quality, generalization ability, and adaptability to real-world scenarios. Code is available at https://github.com/dusongcheng/UniFusion.
\end{abstract}

\section{Introduction}
\label{sec:intro}
Image fusion seeks to integrate complementary information from multiple source images into a single, informative, and visually coherent representation~\cite{zhao2023cddfuse,zou2026toward}. As a fundamental problem in computer vision, it plays a critical role in enhancing scene understanding, facilitating robust decision-making, and supporting downstream tasks such as object detection~\cite{liu2024coconet}, medical diagnosis~\cite{lan2026performance}, autonomous navigation~\cite{liu2023multi},  and low-light surveillance~\cite{yi2024text}. 

Driven by the rapid progress in deep learning, image fusion has undergone remarkable advances in recent years~\cite{zhao2023spherical,zhao2022discrete, yang2025instruction}. Most methods are primarily designed for specific fusion scenarios, including Infrared and Visible Image Fusion (IVIF)~\cite{zhao2020didfuse,zhao2023metafusion,liu2024promptfusion}, multi-exposure fusion~\cite{gupta2021hdr,ouyang2025fusiongcn}, multi-focus fusion~\cite{xiao2021dtmnet,bouzos2023convolutional}, and medical image fusion~\cite{safari2023medfusiongan,liang2024medical}. These methods employ customized designs and utilize diverse architectures such as CNNs~\cite{li2023progressive}, autoencoders, and GANs to extract features for image fusion~\cite{ma2026fastvmt}. However, these task-specific methods have limited generalization capabilities and struggle to adapt to diverse fusion requirements, prompting researchers to explore general fusion frameworks.

To adapt to diverse tasks, recent general image fusion methods aim to handle multiple heterogeneous inputs and fusion tasks with a single model, exhibiting diverse development trends in recent years. Transformer-based architectures~\cite{ma2022swinfusion,zhao2023cddfuse} leverage self-attention mechanisms to model global dependencies across modalities and process different types of image pairs through unified feature interaction modules~\cite{liu2025dcevo}; diffusion model-based methods~\cite{zhao2023ddfm,huang2024fusiondiff} iteratively denoise to generate high-quality fusion results through universal data distributions; other methods~\cite{xu2020u2fusion,yi2024text} construct unified representation spaces by designing task-agnostic fusion modules and shared feature extractors.

Although these methods have achieved significant breakthroughs in fusion quality and semantic understanding, the generalization capabilities of existing general fusion systems remain limited. These approaches generally employ shared backbones and task-agnostic fusion modules to build a common representation space. While promising, their generalization capability is still constrained by two key factors: (1) the lack of a principled, modality-consistent feature extraction mechanism that can robustly encode heterogeneous signals under diverse conditions, and (2) the difficulty in preserving critical information from each source image during deep feature propagation, which leads to information degradation and sub-optimal fusion quality across tasks.

To address these limitations, we propose UniFusion, a robust and unified image fusion framework that enhances the understanding and preservation of source information while achieving superior cross-task generalization. First, we incorporate the powerful self-supervised visual representation capability of DINOv3~\cite{simeoni2025dinov3} to construct a generalizable feature extraction backbone, enabling modality-consistent and robust feature learning across diverse image types. Second, we introduce a reconstruction-alignment mechanism that leverages intermediate features to reconstruct source images, enforcing strong information retention and ensuring faithful semantic alignment during fusion. Finally, we formulate fusion and reconstruction as a bilevel optimization problem, decoupling and jointly optimizing the two objectives to achieve an improved balance between information preservation and fusion quality. Comprehensive experiments across multi-modal, multi-exposure, and multi-focus tasks validate the effectiveness and generalization ability of UniFusion. Our main contributions are summarized as follows:
\begin{itemize}
\item[$\bullet$] A unified fusion architecture with strong cross-task generalization. We present a universal fusion framework that effectively addresses the generalization limitations of existing fusion models.
\item[$\bullet$] Modality consistent feature extraction and reconstruction aligned fusion. We design a universal feature extractor based on DINOv3 and introduce a reconstruction-alignment mechanism that enforces consistent understanding and preservation of source information.
\item[$\bullet$] Bilevel optimization for balanced fusion and source fidelity. We formulate fusion and reconstruction as a bilevel optimization problem, achieving superior fusion performance while better retaining structural, semantic, and modality-specific information from source images.
\end{itemize}
\section{Related work}
\label{sec:Related work}

\subsection{Task-specific Image Fusion Methods}

Image fusion is a fundamental enhancement technique and a long-standing research focus~\cite{zou2026contourlet,ma2025controllable}. Within this field, mainstream schemes, particularly task-specific ones, can be classified into the following categories.

\noindent{\textbf{Infrared-visible Image Fusion}}: This task aims to integrate thermal information from infrared images~\cite{li2025difiisr} with texture details from visible images~\cite{li2025mulfs}. Early deep learning approaches include supervised convolutional neural networks~\cite{liu2022target}. To mitigate misalignment, generation-registration-fusion pipelines~\cite{wang2022unsupervised} and spatial-frequency integration networks~\cite{zhou2024general,du2026unsupervised} have been developed.

\noindent{\textbf{Medical Image Fusion}}: This task focuses on preserving distinct tissue characteristics from multi-modal medical data (e.g., CT/MRI) to support clinical diagnosis, with a focus on content-aware feature preservation~\cite{gong2025med,lan2026reco}. Das et al. proposed a content-aware GAN~\cite{das2024end} for this purpose. Tang et al. developed FATFusion~\cite{tang2024fatfusion}, a dual-branch Transformer framework that explicitly models the distinct properties of functional and anatomical modalities.

\noindent{\textbf{Multi-exposure Image Fusion}}: This task focuses on synthesizing a high-dynamic-range image from a sequence of captures of the same scene under varying exposures~\cite{fangphoton,li2024contourlet}. The core challenge lies in achieving balanced brightness and preserving fine details. Xu et al. proposed MEF-GAN~\cite{xu2020mef}, and Zhang et al. further optimized this with MFF-GAN~\cite{zhang2021mff}. 
Ouyang et al.\cite{ouyang2025fusiongcn} combine GCNs and lightweight CNNs for real-time applications.

\subsection{General Image Fusion Methods}

While task-specific methods demonstrate strong performance within their domains by leveraging specialized priors, they often operate in isolation, overlooking the potential synergies and common underlying principles across different fusion tasks. This limitation has motivated the development of unified frameworks that aim to capture these inherent associations.

\noindent{\textbf{Advanced Architecture-based Methods}}: This line of research focuses on building unified frameworks by leveraging the complementary strengths of CNNs and Transformers~\cite{zhao2024equivariant,fang2025integrating}. Liu et al. proposed CSR \cite{liu2016image} using a convolutional sparse representation. Subsequently, U2Fusion pioneered an all-in-one approach~\cite{xu2020u2fusion}. Ma et al. developed SwinFusion \cite{ma2022swinfusion}, a cross-domain Swin Transformer framework. 
Duan et al. designed a hybrid CNN-Transformer architecture \cite{duan2024combining} with online knowledge distillation.

\noindent{\textbf{Large Model and Emerging Paradigm-based Methods}}: Recent trends explore integrating capabilities from large models~\cite{ma2025followyourclick,wei2026dynamicgtr,li2026toward,li2025mmt}, such as text guidance~\cite{li2023text,xu2025clip,shen2025fine}, diffusion models~\cite{wangefficient}, and adapter modules~\cite{shi2024vdmufusion}, to enhance semantic awareness and handle cross-task disparities.  Building upon these advances, Yi et al. introduced Text-IF~\cite{yi2024text}, a text-guided framework. Zhu et al. proposed TC-MoA~\cite{zhu2024task}, which uses a task-specific routing network to unify multiple fusion tasks.
\section{Method}

\begin{figure*}
    \centering
    \includegraphics[width=1\linewidth]{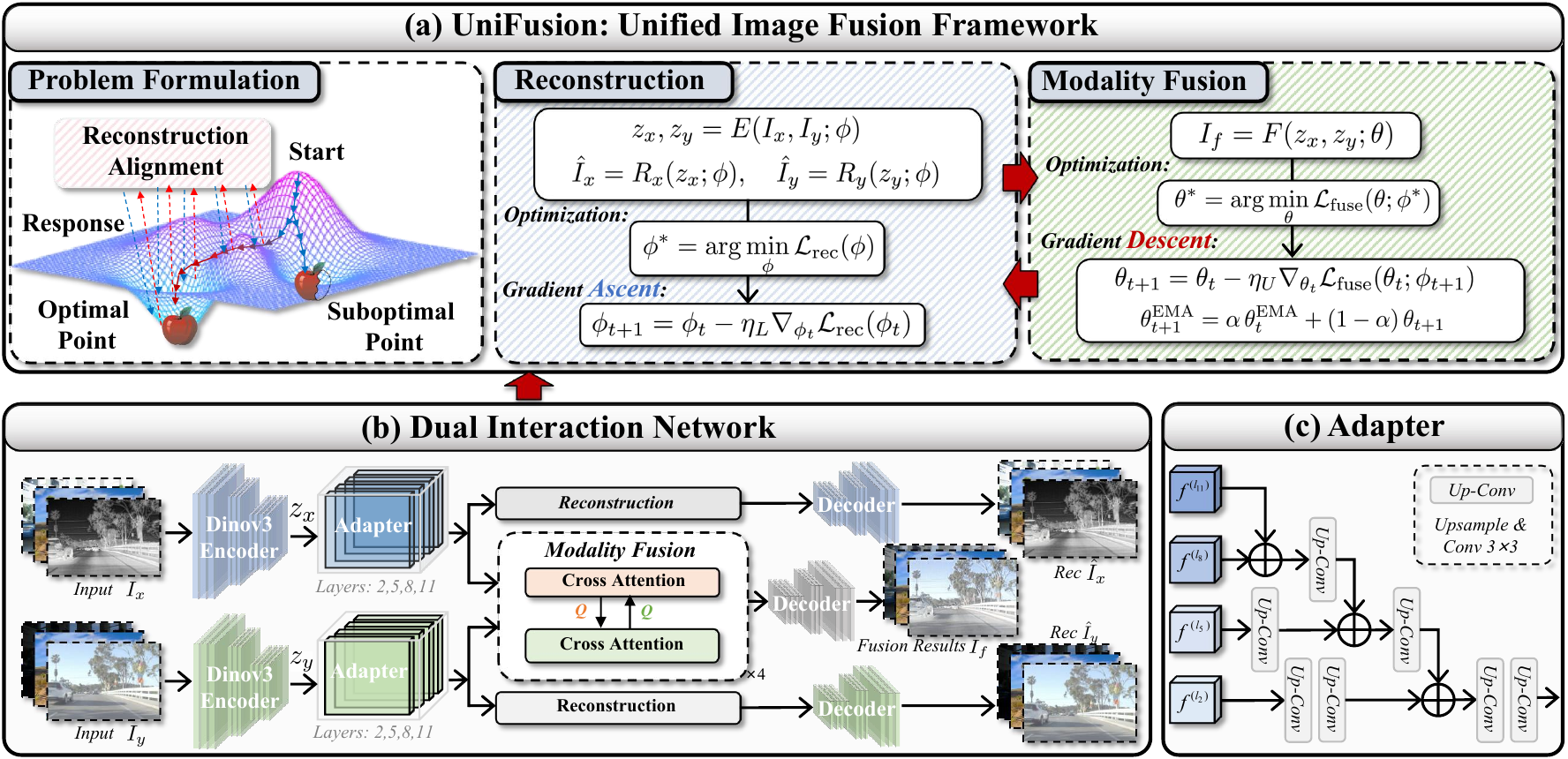}
    \caption{Overview of the proposed UniFusion framework.}
    \label{fig:framework}
\end{figure*}

\subsection{Overview}
We propose a unified image fusion framework that leverages the semantic representation capability of DINOv3 to achieve modality-consistent and generalizable feature integration. As illustrated in ~\cref{fig:framework}, two frozen DINOv3 backbones are employed to extract semantically rich features from each modality, while lightweight adapters are introduced to modulate domain-specific characteristics and reduce distributional discrepancies. The adapted features are then fused through several cross-attention modules that dynamically model inter-modality dependencies and emphasize complementary information. To avoid information loss and semantic drift, each adapter is paired with a reconstruction branch that decodes its adapted embedding back to the original input and is supervised independently. Finally, the entire framework is jointly optimized under a hybrid objective that balances semantic alignment, reconstruction fidelity, and fusion consistency, ensuring both effective information preservation and semantic-level integration across diverse fusion scenarios.

\subsection{Semantic Prior Adaptation with DINOv3}
To achieve modality-consistent and generalizable feature representation, we employ DINOv3, a self-supervised Vision Transformer pretrained on large-scale natural images, as the universal semantic backbone. Unlike conventional modality-specific encoders that are prone to overfitting and fail to generalize across domains, DINOv3 offers object-centric priors and long-range contextual dependencies, which provide a strong foundation for cross-modal representation learning. Given an input image \(I_m \in \mathbb{R}^{H \times W \times 3}\) from modality \(m\), the image is first divided into non-overlapping patches and linearly projected into \(d\)-dimensional token embeddings. These tokens are propagated through \(L\) transformer layers of the frozen DINOv3 encoder, from which a subset of intermediate feature maps is extracted to form a semantic hierarchy that jointly encodes local structural cues and global contextual semantics.

Despite its strong generalization, DINOv3 is pretrained on large-scale natural images, and its latent space may not perfectly align with modality-specific characteristics. To address this domain discrepancy, we introduce a lightweight, hierarchical Adapter that performs progressive feature recalibration across multiple semantic scales.  

Conceptually, the Adapter functions as a progressive feature translator, refining multi-level ViT representations into modality-aligned embeddings. It employs a cascade of multi-stage fusion and upsampling operations, in which transformer features extracted from multiple layers (\(f^{(l_2)},\ f^{(l_5)},\ f^{(l_8)},\ f^{(l_{11})}\)) are sequentially integrated through a residual hierarchy. Each fusion stage aggregates global semantic cues from deeper layers with fine-grained structural details from shallower ones, enabling a smooth semantic–structural transition across scales. This hierarchical integration not only enhances representational fidelity but also enforces distributional consistency across modalities, providing a robust and unified feature space for subsequent fusion and reconstruction.

\subsection{Reconstruction Alignment}
Although the proposed dual-branch architecture effectively adapts pretrained DINOv3 representations to diverse modalities, direct fusion of such embeddings may still suffer from semantic drift and information loss. As features from heterogeneous sources are projected into a shared latent space, modality-specific cues, such as texture in visible images or radiometric contrast in infrared ones, tend to weaken, leading to suboptimal alignment and degraded reconstruction fidelity.

Existing image fusion methods predominantly emphasize the similarity between the fused output and the source images, often adopting pixel-level or appearance-based supervision to preserve visual details. However, such low-level constraints tend to bias the model toward reproducing superficial texture or intensity patterns, rather than capturing deeper semantic correspondences across modalities. Consequently,  these often overfit to appearance details while failing to maintain high-level consistency. 

To mitigate this issue, we introduce a reconstruction-alignment mechanism that explicitly regularizes the fusion process through self-reconstruction constraints. The core idea is to ensure that the encoded latent representation retains sufficient modality-specific information to reconstruct the original inputs, thereby enforcing semantic fidelity and preventing information loss.

Concretely, for each modality \(m\), the adapter produces a calibrated feature map \(\hat{\mathbf{F}}_m\), which is passed through a lightweight reconstruction branch composed of a few Transformer layers and a projection head:
\begin{equation}
\bar{I}_m = R_m(\hat{\mathbf{F}}_m)
\end{equation}
where \(R_m(\cdot)\) denotes the reconstruction function.

\subsection{Bilevel Optimization}
To jointly optimize semantic reconstruction and modality fusion operating at different temporal dynamics, we formulate the learning process as a bilevel optimization problem. In this hierarchical formulation, the inner loop rapidly updates reconstruction-oriented representations to capture modality-specific semantics, while the outer loop gradually adjusts the fusion strategy based on the evolving feature space. This two-layer interaction facilitates stable co-adaptation between feature encoding and fusion, ultimately achieving a balanced tradeoff between semantic consistency and structural fidelity in the final fused result.

Formally, let \(\phi\) denote the parameters of the adapter and reconstruction subnetworks responsible for modality-specific representation, and \(\theta\) represent the parameters of the fusion network. Given an image pair (\(I_x\), \(I_y\)), the feature encoding and reconstruction processes can be expressed as:
\begin{equation}
    z_{x}, z_{y} = E(I_{x}, I_{y}; \phi), \quad
\hat{I}_{x} = R_{x}(z_{x}; \phi), \quad
\hat{I}_{y} = R_{y}(z_{y}; \phi),
\end{equation}
while the fusion result is obtained by:
\begin{equation}
    I_f = F(z_x, z_y; \theta).
\end{equation}
The overall training objective is then formulated as a bilevel optimization problem:
\begin{equation}
    \phi^* = \arg\min_{\phi} \mathcal{L}_{\mathrm{rec}}(\phi),
    \theta^* = \arg\min_{\theta} \mathcal{L}_{\mathrm{fuse}}(\theta; \phi^*)
\end{equation}

The inner-level objective focuses on reconstructing each modality to preserve semantic and structural fidelity, guiding the encoders to produce features that are rich in modality-aware cues. The outer-level objective then learns an optimal fusion strategy based on these representations, ensuring complementary information is integrated effectively.

In practice, directly solving this nested optimization is computationally expensive. Therefore, we employ a first-order alternating scheme, where the two parameter sets are updated sequentially within each iteration:
\begin{equation}
\begin{aligned}
    \phi_{t+1} &= \phi_t - \eta_L \nabla_{\phi_t} \mathcal{L}_{\mathrm{rec}}(\phi_t), \\
    \theta_{t+1} &= \theta_t - \eta_U \nabla_{\theta_t} \mathcal{L}_{\mathrm{fuse}}(\theta_t; \phi_{t+1}),
\end{aligned}
\end{equation}
where \(\eta_L\) and \(\eta_U\) denote the learning rates for the inner and outer levels, respectively, typically satisfying \(\eta_L > \eta_U\) to ensure faster representation adaptation.

To further enhance the temporal stability of the fusion dynamics, we introduce an exponential moving average (EMA) regularization on the outer parameters:
\begin{equation}
    \theta_{t+1}^{\mathrm{EMA}} = \alpha \, \theta_t^{\mathrm{EMA}} + (1 - \alpha) \, \theta_{t+1},
\end{equation}
where \(\alpha \in [0,1)\) controls the momentum of the update. 

\begin{table*}
\centering
\resizebox{\linewidth}{!}{\begin{tabular}{l|cccc|cccc|cccc}
\toprule[1.2pt]
Dataset    & \multicolumn{4}{c|}{M\(^3\)FD Dataset}     & \multicolumn{4}{c|}{T\&R Dataset}         & \multicolumn{4}{c}{MEFB Dataset}                                            \\
\midrule
           Method & MI\(\uparrow\) & VIF\(\uparrow\) & \(Q_{abf}\uparrow\)& \(Q_{y}\uparrow\) & MI\(\uparrow\) & VIF\(\uparrow\) & \(Q_{abf}\uparrow\)& \(Q_{y}\uparrow\)  & MI\(\uparrow\) & VIF\(\uparrow\)& CC\(\uparrow\)  & PSNR\(\uparrow\) \\
\midrule
CDDFuse\cite{zhao2023cddfuse}    & \underline{3.776}          & 0.839          & 0.610          & 0.978          & 3.102          & 0.732                & 0.489                & 0.945                & \underline{6.575}                & 1.430                & 0.837                & 56.809\\
CoCoNet~\cite{liu2024coconet}    & 2.621          & 0.721          & 0.378          & 0.826          & 2.576          & 0.602                & 0.362                & 0.781                & 4.607                & 1.091                & 0.858                & 58.007\\
DeFuse~\cite{liang2022fusion}     & 2.924          & 0.612          & 0.332          & 0.959          & 2.991          & 0.667                & 0.393                & \underline{0.968}                & 5.238                & 1.164                & 0.813                & 58.209\\
LRRNet~\cite{li2023lrrnet}     & 2.763          & 0.629          & 0.495          & 0.950          & 2.766          & 0.577                & 0.352                & 0.781                & 5.946                & 1.110                & 0.885                & 58.970\\
MGDN ~\cite{guan2023mutual}      & 2.832          & 0.750          & 0.618          & 0.961          & 2.810          & 0.701                & 0.552                & \underline{0.968}                & 4.935                & 1.148                & 0.838                & 58.516\\
ReCoNet~\cite{huang2022reconet}    & 3.009          & 0.682          & 0.485          & 0.937          & 2.972          & 0.624                & 0.375                & 0.958                & 5.583                & 1.074                & 0.854                & 57.700\\
SwinFusion~\cite{ma2022swinfusion} & 2.945          & 0.618          & 0.480          & 0.936          & \underline{3.325}          & 0.747                & 0.467                & 0.974                & 5.318                & \underline{1.459}                & 0.894                & 59.009\\
U2Fusion~\cite{xu2020u2fusion}   & 2.723          & 0.688          & 0.535          & 0.860          & 2.592          & 0.660                & 0.488                & 0.918                & 5.809                & 1.284                & \underline{0.902}                & 58.890\\
UMFusion~\cite{wang2022unsupervised}   & 3.099          & 0.675          & 0.399          & 0.861          & 2.893          & 0.711                & 0.470                & 0.937                & 6.326                & 1.293                & 0.889                & 58.881\\
TC-MoA~\cite{zhu2024task}     & 3.466          & \underline{0.870}          & \underline{0.636}          & \textbf{0.983} & 3.038          & \underline{0.787}                & \underline{0.558}                & 0.962                & 4.889                & 1.406                & 0.885                & \underline{59.152}\\
\rowcolor[gray]{0.92} Ours       & \textbf{4.268} & \textbf{0.899} & \textbf{0.637} & \underline{0.982}          & \textbf{3.599} & \textbf{0.805}       & \textbf{0.565}       & \textbf{0.980}       & \textbf{6.861}       & \textbf{1.484}       & \textbf{0.906}       & \textbf{59.219} \\
\bottomrule[1.2pt]
\end{tabular}}
\caption{Quantitative comparison on M\(^3\)FD, TNO, RoadScene, and MEFB datasets. The best and second best results are highlighted in \textbf{bold} and \underline{underline}.}
\label{tab:m3fd&T&R}
\end{table*}

\begin{figure*}
    \centering
    \includegraphics[width=1\linewidth]{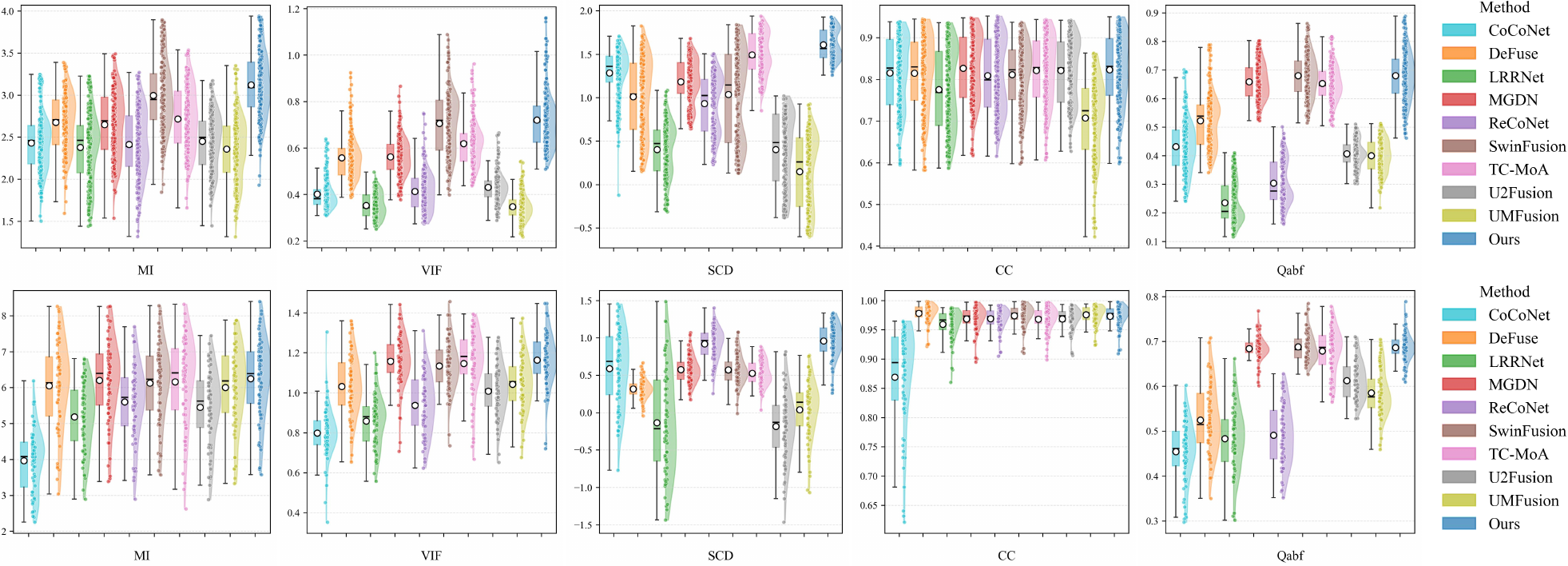}
    \caption{Quantitative comparison on MIF (top) and MFIF (bottom) datasets. The plots illustrate the distribution of all test samples across five evaluation metrics, where “–” and “\(\circ\)” indicate the median and mean values, respectively.}
    \label{fig:mif_mfif}
\end{figure*}

\begin{figure*}
    \centering
    \includegraphics[width=1\linewidth]{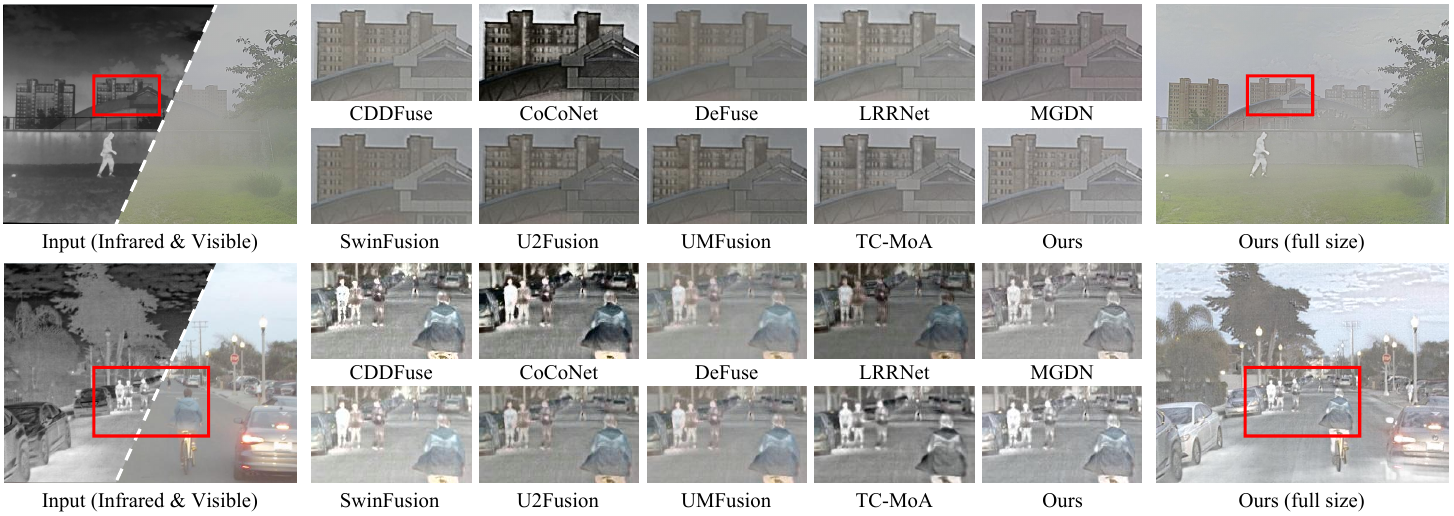}
    \caption{Visual comparison of infrared and visual image fusion results with SOTA methods on M\(^3\)FD (top) and T\(\&\)R (bottom) datasets.}
    \label{fig:show_vif}
\end{figure*}

\begin{figure}
    \centering
    \includegraphics[width=1\linewidth]{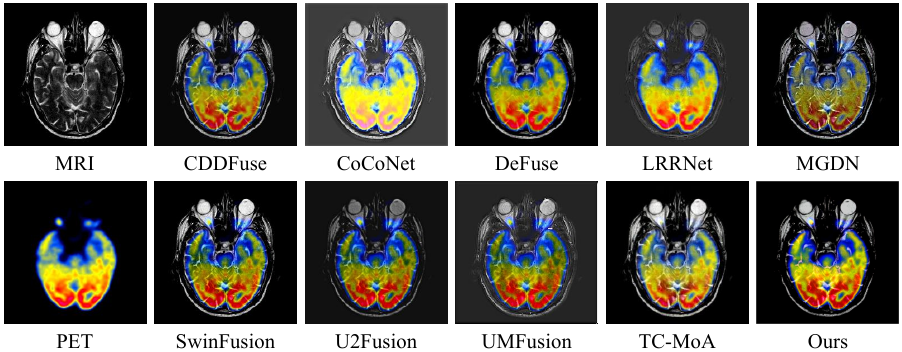}
    \caption{Visual comparison of medical image fusion results with SOTA methods on MIF dataset.}
    \label{fig:show_med}
\end{figure}

\begin{figure}
    \centering
    \includegraphics[width=1\linewidth]{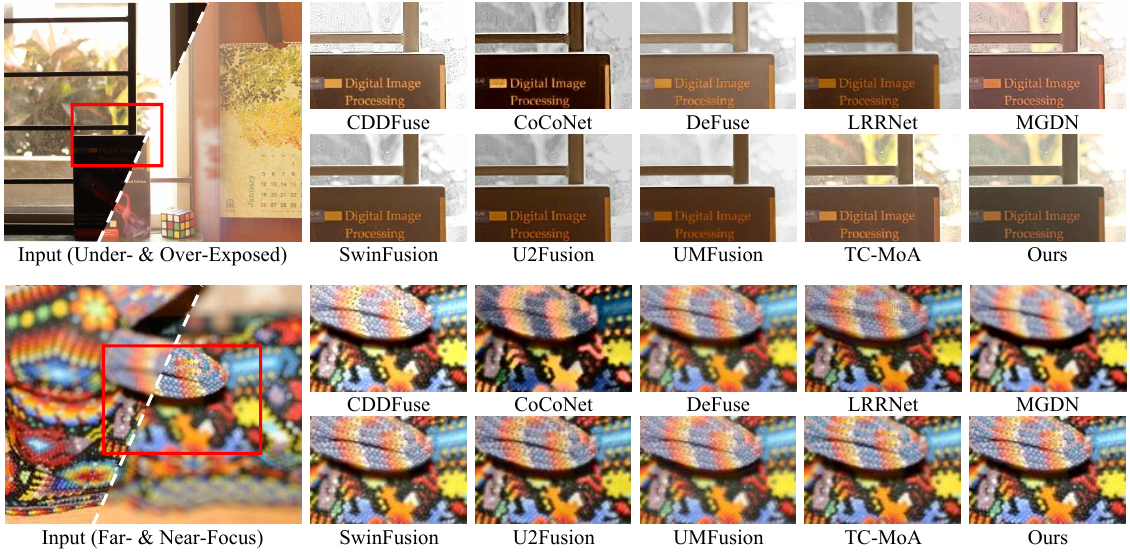}
    \caption{Visual comparison of multi-exposure image fusion and multi-focus image fusion results with SOTA methods on MEFB and MFIF datasets.}
    \label{fig:show_mef_mfif}
\end{figure}

\section{Experiment}
\subsection{Experimental Setting}
\noindent\textbf{Dataset.} We evaluate the proposed framework across four representative image fusion tasks, including infrared–visible image fusion (IVIF), medical image fusion (MIF), multi-exposure image fusion (MEF), and multi-focus image fusion (MFF). For IVIF, experiments are conducted on three widely adopted benchmark datasets: M3FD~\cite{liu2022target}, TNO~\cite{toet2012progress}, and RoadScene~\cite{xu2020fusiondn}, which collectively cover diverse illumination, object, and background conditions. For MIF tasks, we employ the Medical Harvard dataset~\footnote{http://www.med.harvard.edu/AANLIB/home.html}. The MEF and MFF tasks are evaluated using the MEFB~\cite{zhang2021benchmarking} and MFIF~\cite{zhang2021deep} benchmarks, respectively. The MFIF collection integrates multiple sub-datasets, including Lytro~\cite{nejati2015multi}, MFFW~\cite{xu2020mffw}, and MFI-WHU~\cite{zhang2021mff}, offering a comprehensive evaluation of multi-focus fusion under varied scene configurations.

\noindent\textbf{Implementation Details.} Our experiments are performed on an NVIDIA RTX 5090 GPU. The batch size is set to 16, and each fusion task is trained for 10,000 iterations. During training, images are randomly cropped into 128×128 patches and normalized to the range [0,1]~\cite{li2022drcr,shao2025egohiermask}. Model parameters are optimized using the Adam optimizer, with the learning rate initialized to \(2\times10^{-4}\) and decayed exponentially during training~\cite{tcsvt1,guo2026momentum}. The fusion network consists of four Cross-Attention Blocks, while each reconstruction branch includes four Transformer Blocks. The fusion optimization follows the loss formulation of SwinFusion~\cite{ma2022swinfusion}, ensuring fair comparison and stable convergence across different fusion tasks. In addition, we incorporate a reconstruction loss in a self-supervised manner, where each source image is reconstructed from the fused representation using an L1 loss.

\noindent\textbf{Benchmarks.} We compare our method against ten recent state-of-the-art image fusion approaches, including CDDFuse~\cite{zhao2023cddfuse}, CoCoNet~\cite{liu2024coconet}, DeFuse~\cite{liang2022fusion}, LRRNet~\cite{li2023lrrnet}, MGDN~\cite{guan2023mutual}, ReCoNet~\cite{huang2022reconet}, SwinFusion~\cite{ma2022swinfusion}, U2Fusion~\cite{xu2020u2fusion}, UMFusion~\cite{wang2022unsupervised}, and TC-MoA~\cite{zhu2024task}.

\subsection{Evaluation on Multi-Modal Image Fusion}
\noindent\textbf{Quantitative Comparisons.} We quantitatively evaluate the fusion performance using four widely adopted metrics: Mutual Information (MI), Visual Information Fidelity (VIF), \(Q_{abf}\) and \(Q_{y}\), as shown in ~\cref{tab:m3fd&T&R}. Our method consistently outperforms existing general-purpose fusion algorithms across multiple infrared–visible fusion benchmarks, demonstrating strong generalizability and modality compatibility. Notably, the improvements in VIF, \(Q_{abf}\) and \(Q_{y}\), which are closely related to human visual perception and information preservation, indicate that the fused images generated by our framework retain richer structural and semantic details from the source inputs. In addition, the higher MI value further confirms that our method effectively maintains both mutual information characteristics, leading to more perceptually faithful and information-rich fusion results.

\noindent\textbf{Qualitative Comparisons.} 
As illustrated in ~\cref{fig:show_vif}, our method exhibits superior visual performance compared with other approaches. For qualitative analysis, we selected two representative samples from the M3FD and RoadScene datasets, covering diverse scenarios to ensure a comprehensive evaluation. The proposed method more effectively preserves fine texture details from the original images, such as the textures of building windows and the intricate structures at the ends of tree branches, whereas other methods tend to blur these features. Moreover, our approach highlights human contours more clearly and better maintains the mutual information between visible and infrared images. These results demonstrate the clear advantages of our method in terms of qualitative performance.

Additional results on downstream perception tasks, including object detection and semantic segmentation~\cite{li2025stitchfusion}, are provided in the Appendix to further validate the effectiveness of our fusion method.

\subsection{Evaluation on Medical Image Fusion}
\noindent\textbf{Quantitative Comparisons.}
We conduct quantitative experiments on the MIF dataset using five standard metrics, as illustrated in ~\cref{fig:mif_mfif}. The results visualize the distribution of all test samples, along with their mean and median values, revealing stable improvements across most indicators. The consistently high median scores indicate that our method maintains reliable fusion quality across diverse clinical cases, validating its robustness and effectiveness in preserving complementary modality information.


\noindent\textbf{Qualitative Comparisons.} 
The visual comparisons of PET–MRI fusion results are presented in ~\cref{fig:show_med}. Competing methods often blur anatomical structures or lose soft-tissue details when PET lacks functional content, reflecting insufficient semantic alignment. In contrast, our method preserves the rich structural information from MRI while accurately integrating PET functional cues. This benefit arises from the DINOv3-guided semantic priors and hierarchical adapter design, which enable precise cross-modality calibration. Consequently, the fused images achieve both structural clarity and functional expressiveness, supporting more reliable medical interpretation.

\subsection{Evaluation on Multi-Exposure Fusion}
\noindent\textbf{Quantitative Comparisons.}
We quantitatively evaluate the multi-exposure image fusion performance using four widely adopted metrics: Mutual Information (MI), Visual Information Fidelity (VIF), CC, and PSNR, as shown in ~\cref{tab:m3fd&T&R}. Our model achieves the SOTA results among general image fusion approaches, consistently outperforming competing models across most metrics. In particular, the substantial improvement in MI and VIF scores highlights the strong task adaptability of UniFusion, which effectively reconciles task-specific objectives while mitigating potential conflicts between them. Moreover, the higher CC and PSNR scores demonstrate that our method preserves structural fidelity, fine-grained textures, and overall perceptual quality more effectively.

\noindent\textbf{Qualitative Comparisons.} 
As illustrated in ~\cref{fig:show_mef_mfif}, our method consistently produces visually superior fusion results compared with existing SOTA approaches. It preserves fine texture details and maintains accurate color consistency across varying illumination levels, resulting in clearer and more balanced fused images. These results further substantiate the advantages of our method in producing visually coherent and information-rich outputs.

\begin{figure*}
    \centering
    \includegraphics[width=1\linewidth]{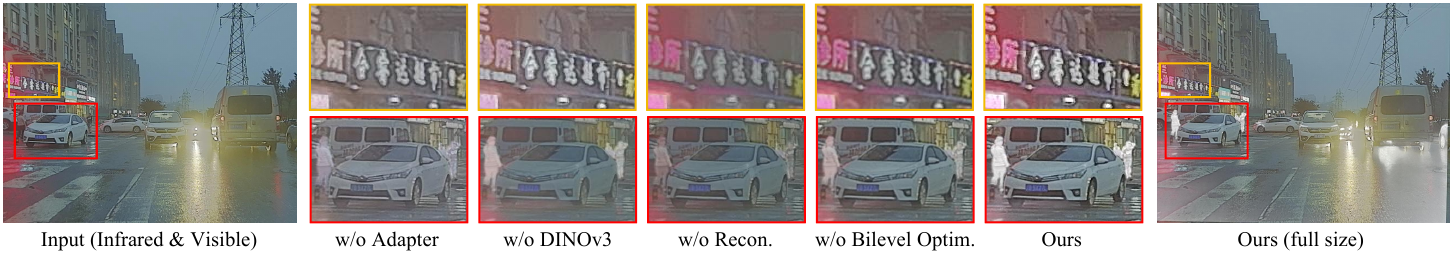}
    \caption{Visual comparison of fusion results for different ablation variants on representative samples from the M\(^3\)FD dataset.}
    \label{fig:ablation}
\end{figure*}

\subsection{Evaluation on Multi-Focus Fusion}
\noindent\textbf{Quantitative Comparisons.}
We further assess UniFusion on the MFIF dataset to test its generalization to the multi-focus fusion (MFF) task. As shown in ~\cref{fig:mif_mfif}, our method consistently ranks among the top two across all five metrics, even without task-specific fine-tuning. This demonstrates its strong adaptability to unseen domains and effectiveness in preserving salient details across multiple focal depths.

\noindent\textbf{Qualitative Comparisons.} 
Representative examples from the MFIF dataset are shown in ~\cref{fig:show_mef_mfif}. Our results exhibit sharper transitions and more consistent focus across regions, retaining fine textures in both near- and far-focus areas. In contrast, TC-MoA shows blur in distant regions, while LRRNet distorts near-focus details. These results highlight the superior perceptual quality and balanced fusion achieved by our method.

\begin{figure}
    \centering
    \includegraphics[width=1\linewidth]{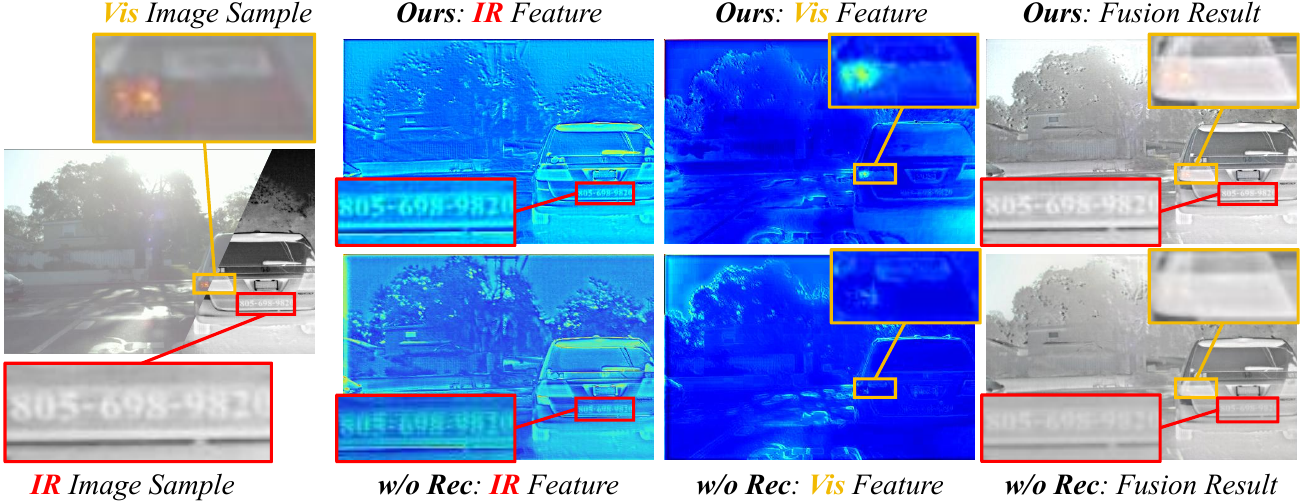}
    \caption{Visualization of encoded features and fusion results with and without the reconstruction alignment module  (“Ours” vs “w/o Rec”). The feature maps are extracted from the encoders for the visible and infrared modalities, respectively.}
    \label{fig:ablation_rec}
\end{figure}

\begin{table*}
\centering
\resizebox{\linewidth}{!}{\begin{tabular}{l|cccc|cccc|cccc}
\toprule[1.2pt]
Dataset    & \multicolumn{4}{c|}{M\(^3\)FD Dataset}     & \multicolumn{4}{c|}{MEFB Dataset}         & \multicolumn{4}{c}{MFIF Dataset}                                            \\
\midrule
           Configurations & MI\(\uparrow\) & VIF\(\uparrow\) & \(Q_{abf}\uparrow\)& \(Q_{y}\uparrow\) & MI\(\uparrow\) & VIF\(\uparrow\)& CC\(\uparrow\)  & PSNR\(\uparrow\) & MI\(\uparrow\) & VIF\(\uparrow\) & CC\(\uparrow\) & \(Q_{abf}\uparrow\)  \\
\midrule
w/o Adapter              & 3.646  & 0.863  & 0.520  & 0.911      & 5.512   & 1.232   & 0.489   & 57.934     & 5.375    & 0.969    & 0.959    & 0.532 \\
w/o DINOv3 encoder       & 3.681  & 0.879  & 0.602  & 0.946      & 5.709   & 1.334   & 0.362   & 58.537     & 5.624    & 0.901    & 0.947    & 0.491 \\
w/o Reconstruction       & 3.846  & 0.870  & 0.587  & 0.932      & 6.434   & 1.396   & 0.393   & 58.779     & 5.838    & 1.013    & 0.961    & 0.579 \\
w/o Bilevel Optimization & 3.924  & 0.876  & 0.621  & 0.959      & 6.374   & 1.424   & 0.393   & 58.960     & 6.021    & 1.032    & 0.964    & 0.583 \\
\rowcolor[gray]{0.92} Ours                     & \textbf{4.268}       & \textbf{0.899}       & \textbf{0.637}       & \textbf{0.982}   
                         & \textbf{6.861}       & \textbf{1.484}       & \textbf{0.906}       & \textbf{59.219} 
                         & \textbf{6.253}       & \textbf{1.159}       & \textbf{0.973}       & \textbf{0.685}  \\
\bottomrule[1.2pt]
\end{tabular}}
\caption{Ablation studies on M\(^3\)FD, MEFB and MFIF datasets. The best and second best results are highlighted in \textbf{bold}.}
\label{tab:ablation}
\end{table*}

\subsection{Ablation Study}
To evaluate the effectiveness and necessity of each core component in the proposed framework, we conducted a series of ablation experiments on three representative datasets, namely M3FD, MEFB, and MFIF. As summarized in ~\cref{tab:ablation}, we designed four carefully constructed model variants to isolate and assess the contributions of key modules and training strategies. In addition, we further provide qualitative visualizations of representative samples to intuitively illustrate the influence of each module on fusion quality. As shown in ~\cref{fig:ablation}, the full model produces clearer structural details, richer textures, and more natural visual effects compared to the ablated variants, confirming the effectiveness of the proposed components in enhancing both perceptual fidelity and information preservation.

\noindent\textbf{w/o Adapter.}
In this variant, the two modality-specific adapters were removed and replaced with simple upsampling operations for feature alignment. Without the adapter’s ability to dynamically modulate modality-specific information, the model struggled to balance two input modality representations. This resulted in a noticeable decline in overall fusion quality, particularly in measures related to information preservation and perceptual consistency. The results suggest that the adapter serves as an essential bridge between heterogeneous modalities, facilitating more coherent spatial feature integration.

\noindent\textbf{w/o DINOv3 Encoder.}
To examine the role of the pretrained semantic encoder, the DINOv3 encoder was replaced with a standard four-layer Transformer encoder. A clear degradation was observed across all datasets, especially in metrics reflecting semantic fidelity and perceptual sharpness (e.g., MI, VIF, and \(Q_{abf}\)). These results indicate that the DINOv3 encoder introduces rich high-level semantic priors, which guide the model in aligning features from different modalities and improving the overall perceptual realism of the fused images. 

\noindent\textbf{w/o Reconstruction.}
In this configuration, the reconstruction branches for both modalities were removed, and the network was trained solely with the fusion objective. Without reconstruction supervision, the model tended to overfit to shared content and neglect modality-specific information, leading to reduced texture clarity and weaker cross-modality complementarity. This demonstrates that reconstruction plays a crucial role in retaining modality-dependent details and maintaining balanced feature fusion. 

To further verify the effectiveness of this module, we visualize the encoded features from the encoders for both the visible and infrared modalities, as well as the final fusion outputs. As shown in ~\cref{fig:ablation_rec}, without the reconstruction alignment, the encoded features lose part of their modality-specific semantic representations, such as structural and color-related information. In contrast, our full model (Ours) maintains more faithful modality characteristics in the encoded features, which translates into clearer details and more consistent semantic preservation in the fused results. 
These results confirm that the reconstruction alignment module effectively regularizes feature learning, encouraging balanced preservation of modality-dependent and shared information, and thus plays a pivotal role in achieving high-quality fusion.

\noindent\textbf{w/o Bilevel Optimization.}
Finally, we removed the bilevel optimization strategy. The performance declined consistently across all datasets, with more pronounced drops in metrics associated with information richness (MI metric) and visual quality (VIF metric). This confirms that bilevel optimization effectively decouples the reconstruction and fusion objectives, stabilizing the training process and enabling more reliable feature disentanglement.

\section{Conclusion}
\noindent We propose UniFusion, a unified image fusion framework that achieves superior cross-task generalization while preserving source information fidelity. Unlike existing fusion methods that only focus on combining complementary features, we propose a reconstruction-alignment mechanism that ensures consistent understanding and preservation of source information, allowing modality-specific cues to be retained and aligned throughout the fusion process. Combined with the self-supervised semantic priors of DINOv3, a DINOv3-based feature extractor, and bilevel optimization, it achieves co-adaptation between representation learning and fusion strategy, effectively balancing the trade-off between fusion quality and source information fidelity. Experiments across multi-modal, multi-exposure, and multi-focus tasks show that UniFusion achieves state-of-the-art \mbox{performance}.

{
    \small
    \bibliographystyle{ieeenat_fullname}
    \bibliography{main}
}


\end{document}